\documentclass[10pt,twocolumn,letterpaper]{article}

\usepackage{cvpr}
\usepackage{times}
\usepackage{epsfig}
\usepackage{graphicx}
\usepackage{amsmath}
\usepackage{amssymb}
\usepackage{algorithm}
\usepackage{multirow}
\usepackage[noend]{algpseudocode}
\usepackage{lipsum}
\let\OLDthebibliography\thebibliography
\renewcommand\thebibliography[1]{
\OLDthebibliography{#1}
\setlength{\parskip}{0pt}
\setlength{\itemsep}{0pt plus 0.3ex}
}

\usepackage[breaklinks=true,bookmarks=false]{hyperref}

\cvprfinalcopy 

\def\cvprPaperID{****} 

\begin{document}
\title{Imbalanced Data Learning by Minority Class Augmentation using Capsule Adversarial Networks}
\author{Pourya Shamsolmoali$^a$  , Masoumeh Zareapoor$ ^a$, Linlin Shen$^b$, Abdul Hamid Sadka$^c$, Jie Yang $^{a}$ \thanks{Correspondence to: Jie Yang $  ( jieyang@sjtu.edu.cn ) $.}\\
\and
$^{a}$ {\small Shanghai Jiao Tong University}\\ 
\and
$^{b}$ {\small Institute of Computer Vision, Shenzhen University, Shenzhen, China}\\
\and
$^{c}$ {\small Digital Science and Technology Hub, Brunel University, London, United Kingdom} \\
}
\maketitle
\footnotetext{ Manuscript submitted May 11, 2019; Accepted February 20, 2020.}
\begin{abstract}
The fact that image datasets are often imbalanced poses an intense challenge for deep learning techniques. In this paper, we propose a method to restore the balance in imbalanced images, by coalescing two concurrent methods, generative adversarial networks (GANs) and capsule network. In our model, generative and discriminative networks play a novel competitive game, in which the generator generates samples towards specific classes from multivariate probabilities distribution. The discriminator of our model is designed in a way that while recognizing the real and fake samples, it is also requires to assign classes to the inputs. Since GAN approaches require fully observed data during training, when the training samples are imbalanced, the approaches might generate similar samples which leading to data overfitting. This problem is addressed by providing all the available information from both the class components jointly in the adversarial training. It improves learning from imbalanced data by incorporating the majority distribution structure in the generation of new minority samples. Furthermore, the generator is trained with feature matching loss function to improve the training convergence. In addition, prevents generation of outliers and does not affect majority class space. The evaluations show the effectiveness of our proposed methodology; in particular, the coalescing of capsule-GAN is effective at recognizing highly overlapping classes with much fewer parameters compared with the convolutional-GAN. 
\end{abstract}

\section{Introduction}
Imbalanced data is a common problem in real-world datasets encountered in a variety of applications such as medical diagnosis, information retrieval systems, fraud detection and land cover classification in remote sensing images. The amount of data is increasing day by day which raises the demands for accurate learning systems to classify and analyze data effectively. Learning from imbalanced data cases severe difficulty for accurate classification. The machine learning classifiers do not perform well on imbalanced data and mostly require the larger class samples. However, for deep learning approaches, the issues will be more complicated because the representation learning degrades the performance of the majority class samples [6,16]. The common strategy to handle imbalanced datasets is to resample the data before training and that includes over-sampling, under-sampling, or combination [3,4,21]. \\
\begin{figure}
\centering
 \includegraphics[width=0.5\textwidth]{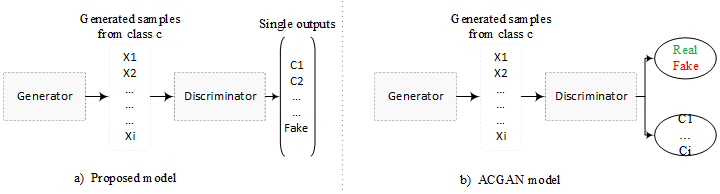}
\caption{Our proposed capsule-discriminator (DC) via ACGAN [20].}
\label{fig:2}       
\end{figure}

\indent {\bf Generative Adversarial Networks (GAN)} [11] is a deep learning method that builds up several layers of abstraction in order to generate real-looking images. Typically, a GAN model is composed of two neural networks, a generator that generates synthetic samples (fake) and a discriminator that decides whether the given samples are real or fake. Both networks are trained simultaneously by playing an adversarial game. The generator learns from the feedback of the discriminator and does not have access to the real data. Over the last few years, a large number of GAN methods used in a variety of applications. Most of these works are based on computer vision problems and involve image generation and image-to-image translation [13,14,16,26]. One pioneering work is DCGAN which was proposed by Radford et al. [15]; it comprises a set of architectures based on CNNs that have since been widely used in many GAN approaches. CNNs are able to extract high-level features from lower level features via convolutions and pooling. However, CNNs have limitations in understanding the complex relationships between lower-level and higher-level features. This relational issue causes a loss of information that may be crucial for the training performance [16]. \par
\begin{figure*}
\centering
 \includegraphics[width=0.78\textwidth]{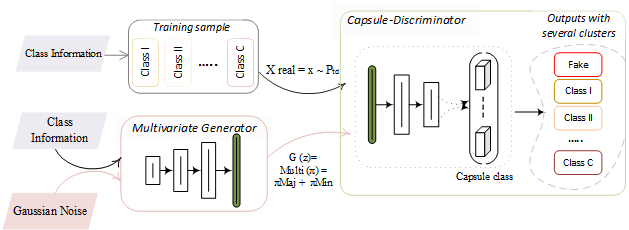}
\caption{The proposed CapsAN architecture. Generated $G_z$ and real data $X_rea$l are fed to the capsule-discriminator (DC). The proposed Multiclass DC creates several clusters in its output space, thus provide a better means for the discriminator that, along with finding generated and real distribution, is also required to predict the class label for the input image. The generator G composed of noise z and class information which uses $Multi (\pi)$ distribution to generate minority class.}
\label{fig:1}       
\end{figure*}
{\bf Capsules} were first introduced in [12] and more recently improved in [18]. This technology offers better performance for learning hierarchical relationships and can be a successful alternative to CNN, especially to change the pose and spatial relationship. The initial intention of deep learning was to design a hierarchical model to recognize the features from low-level to complex ones. However, the capsule network (capsule-net) inherently has this property and can perform more effective than CNN. \\
\indent Nowadays, researchers who work in data imbalance problems mainly focused on developing and modifying the suitable learning classifier to better handle class imbalance [1,7], and less attention is given to the data generation. Despite the remarkable success of GAN in many applications, it is not appropriate for classifying imbalanced data, due to its inherent property which requires fully observed data during training [35,41,42]. We divide the imbalanced datasets into two components, majority class-samples, and minority class-samples. Existing methods in this problem suffer from the following issues [2,39,42,43]:\\
\indent  If the training samples are imbalanced when the generated samples from the minority are passed to the discriminator, they are mostly classified as fake samples. It is due to the fact, with fewer samples; the discriminator cannot find the corresponding class. Secondly, if the generator attempts to generate realistic samples to fool the discriminator, it generally focuses on generating majority class, to optimize its loss function; thus, this will cause the collapse of the model on the minority class. Third, these methods usually interfere with the distribution of the majority class space and leading to slight overfitting of the data. 
To be viable, GANs need to have sufficient information from both class components and define an optimal distribution to tackle these issues. Our proposed Capsule Adversarial Network (CapsAN) has been conceived to address these problems by using both the majority and minority samples in the adversarial training and using a multivariate (mixture) probability distribution. This allows our model to learn the underlying information from the proposed mixture distribution and use them to generate a minority class sample with a higher variety.  To achieve this goal, we need to know the class information in the latent space to generate the samples for the specific class. Although our model learns information from both majority and minority classes, the goal is to generate the samples for the minority class. Moreover, using a single generator to generate samples for imbalance classes may lead to a trivial solution where all the generated samples are similar to the large class component [2,23,41]. To resolve this problem and generate different visual samples, capsule-net is used in the design of our discriminator instead of conventional CNNs.  This combination enables the discriminator to have several embedded clusters in its output (as multiclass discriminator) and prevent overfitting of data. In this way, as there are multiple clusters, the generator is able to generate a high variety of samples from different classes. We prove that using capsule-net as the parameterization of the discriminator has a better performance than CNN-based GAN, particularly for the imbalanced classification task. These comparisons are performed on several imbalanced image datasets. Although some of the existing approaches show promising performance, our models improve upon these methods across various metrics. The proposed model architecture is given in Figure \ref{fig:1}. The rest of the paper is organized as follows: The next section discusses related work~\ref{sec:2}; section~\ref{sec:3} provides a brief discussion about GAN and capsule-net. We present our proposed model in section~\ref{sec:4}. Quantitative and qualitative analysis is shown in section~\ref{sec:5}. Finally, the work is concluded in section~\ref{sec:6}.

\section{Related works}
\label{sec:2}
To address the imbalanced class problems, a lot of work is presented in the literature. A comprehensive comparison of class imbalance and data irregularities problems can be seen in [10,17]. These articles discuss different approaches for data irregularities such as class imbalance, small disjoints, class askew, missing features, and absent features. In [28], the authors marking the fifteen-year anniversary of SMOTE, discuss the present state of affairs with SMOTE, its applications, and also classify the future challenges to extend SMOTE for Big Data problems. \par
Different from the existing GAN-based methods that employ GAN to generate samples, we aim to use GAN for improving imbalanced image classification. However, GAN-based imbalanced image classification is still in its infancy and so far a few works have been proposed. Fiore et al. [27] utilized a GAN to generate minority class samples (fraud samples) for credit card fraud detection. Douzas et al. [29] used GAN to generate synthetic samples for the minority class over various imbalanced datasets and proved that GAN results are higher than those of other oversampling methods. Montahaei et al. [25] proposed ARIC model to handle the imbalanced problems in the adversarial classifiers. Their model is based on weight adjustment factors, in which the generator gives weights to the majority class and the discriminator aims to classify the minority and the weighted majority samples. However, their model fails to consider the minority class distribution; they just uniformly assign the weight to the majority class samples. In addition, their proposed model is computationally complex and expensive. Mullick et al. proposed GAMO [30] which is another method for handling class imbalance based adversarial networks. Their model consists of a generator, a classifier, and a discriminator. Samples are scattered in three distribution structures; majority class, minority class, and generated minority class distribution. The generator of GAMO is trained to generated minority samples that are misclassified from the classifier. The discriminator is trained to find real and fake data instances. The classifier is trained to correctly classify the instances as; majority or minority samples. Their proposed model is ideal and shown a promising performance on different imbalanced datasets. Compared to these techniques, our model improves upon these methods across different metrics, as can be seen in the experiments section. In fact, our model has a different setting and shows a new way to generate minority samples, by incorporating the spatial information of majority class into the newly generated minority class, which is not commonly used in other techniques. The proposed CapsAN is in line with [23] in the sense that we propose a multiclass discriminator, and similar to [32] we used mixture data distribution in order to drive the generation process towards minority class. Unlike the other variation of GANs [30], an additional classifier did not apply within our network; instead, the capsule discriminator of our model requires classifying incoming vectors based on the defined conditions in the generator. 

\section{Convolutional-GANs via Capsule-GANs}
\label{sec:3}
Before delving into the presentation of our proposed model, it is expedient to give a brief description of the material concepts needed for proposing our work. GAN [11] consists of two networks; a generator and a discriminator. The underlying idea is to train a generator against a discriminator in an adversarial mode. We assume ${G: Z\rightarrow X}$ for generator G, where $z$ is the noise space and $X$ is the data space. The discriminator D can be defined as ${D: X\rightarrow [0,1]}$ to distinguish the class of the samples that can be either 0 or 1. The GAN method can be formulated as:
\begin{equation}
  \min_{G}\max_{D}V(D,G)=N_D + N_G              
\label{eq:e1}
\end{equation}\par
${x\in X}$ is from the data distribution ${P_data (x)}$ and ${z\in Z}$ is from the noise distribution ${P_z (z)}$ . During training, $N_G$ is optimized to confuse the $N_D$ by assigning the label 1 (fake) to samples generated from G. However, if the discriminator could identify the instances that come from the generator then the observation is: “the generator produces low-quality data which is distinguishable for discriminator” [13]. The original paper of GAN [18] uses a multilayer perceptron for both the generator and discriminator. Even though this network has a successful performance in a wide range of applications, it is not suitable for imbalanced images. We discuss the challenges of imbalanced datasets in the next section. More recently, Sabour et al. [18] proposed a capsule-net, which is an alternative to CNNs with a strong ability to model the relationships between the image features with much fewer parameters. Similar to convolutional networks, capsules also are a hierarchical image representation that could pass an image through multiple layers of the network. The capsule-net consists of several layers and; each layer is divided into groups of neurons that denoted as capsules [34]. In contrast to a single neuron, a capsule not only learns a good weight for the image classification but also can learn a specific image entity over different viewing conditions such as rotation, lighting, and thickness. Capsules are considered to be active based on the activation of the neurons. However, the neuron activation depends on different features such as pose, position, color, etc, which allow the capsule to capture a specific object in the images and if the object exists in the image, then the capsule is activated. The capsule has the output vectors, where the vector length is determined by the presence of the entity. In the original paper\footnote{\fontfamily{qcr}\selectfont{https://arxiv.org/pdf/1710.09829.pdf }} , the authors used the squashing function to shrink the short vectors to zero and large vectors to 1. Furthermore, to compute the output for each parent (Routing algorithm), the capsule estimates a prediction vector which can be calculated by multiplying its own output by the corresponding weights. In fact, these weights determine the strength of the connection for each capsule. Later, the scalar product between the prediction vector and the output of the possible parent will be computed. The intuitive thought is that, if the obtained scalar product is enough large, then a high coupling coefficient is assigned to that parent, whilst it is decreased for other parents. This mechanism helps to avoid the overlapping problem in the classification task. If we assume, capsule $i$ and $j$ in layer $l$ and the layer above $(l+1)$, in which $u_i$ is the vector output of capsule $i$ and $v_j$ is vector output of capsule $j$ that has $s_j$ input vector, the $v_j$  will be defined as ${v_i = squash(s_j) = \frac{||s_j||^2 s_j}{1+{||s_j||^2||s_j||}}}$. It is worth mentioning that the vector length is limited to being within {\verb|zero and one|}; the short vectors minimum gets to zero length and long vector maximum gets to one (while the orientation of the vectors is preserved). The total input of capsule $j$ is a weighted sum of all vectors ${\hat u_{j|i}}$; and ${\hat u_{j|i}}$ is the prediction vector of capsule $i$ connected to capsule $j$ as: ${\hat u_{i|j}}$ = $W_{ij}v_i$, where the output $v_i$ is a vector of capsule $i$, and $W_{ij}$ is a weight matrix. Further, the total input for capsule $j$ is achieved by: $s_j = \sum_{i}f_{ij}\hat u_{j|i}$ ; $f_{ij}$  is a coupling coefficient and adjusts between capsule $i$ and all the capsules in the above layer $j$, and is determined by the ''\verb|routing-by-agreement|'' technique [18]. In this way, when the prediction vector is similar to the output, the coefficient will be increased, while it is decreased when there is dissimilarity. Based on the advantage of the capsule-net, we propose a new framework for imbalanced image classification that incorporates capsules within the GAN framework. 

\section{Capsule Adversarial Networks}
\label{sec:4}
Synthetic data resampling approaches have achieved a fair amount of success. In this paper, we present a new way to generate minority class samples based on generative adversarial networks. The proposed model consists of a generator and a discriminator. The generator corresponds to an oversampling technique and generating more classification problem to the discriminator. The proposed discriminator acts as a classifier to accurately classify the samples into several clusters. Before going into the details of our proposed model, we discuss the main challenges of GANs in imbalanced datasets. If we train a typical GAN on all available data and generate many random samples, the generated samples are mostly from majority class due to the sufficient number of training samples and also its optimal loss function [22]. On the other side, training GAN with minority classes is not a logic due to the fact that, with the small number of samples the random parameters of our model may generate similar samples and leads to overfitting problems [24,39]. Another way to handle imbalanced data is by down-sampling the training data based on majority class samples. However, random down-sampling will degrade the classification performance, since the selected subset of samples may not be informative enough [25,36]. There is another strategy to train GANs on imbalanced data, jointly using majority and minority class samples. In this method, the generator is able to generate samples from different classes and force the discriminator to believe that the samples are a real image of the corresponding class. Inspired by the work described in [20,36], our generator input is conditioned to draw a target class. We apply class information on the latent space. The generated samples should be meeting the following essential conditions; firstly, be indistinguishable from the original ones, in view of the discriminator; secondly, the discriminator should associate it with the class c. The third condition implies that, although our algorithm uses the distribution of majority samples for the generation of new minority class samples, it does not affect the learning from majority class space. However, there is some conflict between the first and second conditions of the minority class samples. For example, when the generated samples- from minority class- pass to the discriminator, it is mostly assigned as fake samples, which is due to lack of training minority samples. We resolve this problem by designing the discriminator with a multiple cluster as; either fake or belongs to the class c; instead of having binary outputs as in [20,25,29,30]. In this way, we avoid the above conflict and improve the discriminative accuracy. Figure \ref{fig:2} shows the different sets of our discriminator and other relative models such as [20,39]. Moreover, to make our discriminative model as effective as possible, and avoid overfitting of the data, we incorporate the capsule-net in our proposed setting to stabilize the training process by guiding the generator to produce better samples. We followed the original paper [18] with a minor variation to implement the discriminator. The main difference is in the final layer, which in our case contains $1\times16-dimensional capsules$ and for the original paper about $10\times16-dimensional capsules$. 

\subsection{Generator based oversampling}
Generating samples of imbalanced images from a latent vector $z$ is not simple, and we should find an accurate distribution between the class samples to prevent the overfitting problem. In this paper, we propose a multivariate probability distribution from a mixture of majority and minority data distribution in the adversarial training. The proposed generative model is capable to generate a high variety of samples and copes with overfitting of the data. Let $\hat x_{(s+)}$ and $\hat x_{(s-)}$ be images drawn from the minority and majority class samples respectively from the real distribution of ${p_{data} (x)}$. Let G be the generative model of our GAN with $p_g (x)$ distribution, which maps a random vector $z$ to images that have the same support as $x$. The training dataset is divided into majority and minority class sample components. Our central idea is to learn these components jointly in order to approximate the data distribution for the minority samples. We consider the density of data points of random class $c$, as $p_c (f)$ with a threshold $\delta_c$, and $f_c$ be the subset of data on the condition,
$f_c = {f:p_c (k)>\delta_c}$; $\delta_c$ is positive ($> 0$), such that the $f_c$ disjoints with the boundary margin if, ${(\delta_c)}_{c=1}^c$. In practice, it is difficult to access the minority sample distribution, because there are fewer samples in the training set. According to [36] using mixture distribution for the minority class samples will be in the form of $\alpha f_{(s+)}+(1-\alpha)f_c $ ; where $\alpha \in[0,1]$. For a finite training set \verb|{s-,s+,G}|, where the generated samples are indicated by $p_{(s-)}$ and $p_{(s+)}$, if $p_g (x)\ne p_{data} (x)$, it means that, the generator distribution is different from the real data distribution, then it can be considered as mixture data distribution. Based on the above observation, we can consider function $f$ as our proposed generator and thus we formulate the multivariate distribution for the generative model as discussed in [32]: 
\begin{equation}
\begin{aligned}
p_g(x) = \pi p_{s+}(x)+(1-\pi)p_{data}(x) \\
D_g(x) = \frac{\pi p_{s+}(x)+(1-\pi)p_{data}(x)}{p_{data}(x)}\\
=\frac{\pi p_{s+}(x)}{p_{data}(x)}+(1-\pi)
\end{aligned}
\label{eq:2}
\end{equation}
For data imbalanced classification we would like the generator to drive the generation process towards minority class. Furthermore, we model a probability distance in the latent space as: 
$T_c = T(\mu_c, \sum_c)$, where it contains mean vector $\mu_c$ and covariance matrix $\sum_c$. This equation should stand for each class $c$, and we compute $\mu_c$ and $\sum_c$ for both majority and minority classes and then draw a latent vector $Z_c$ from distribution $T_c$. The fake samples are also generated by generator G which uses inputs latent vectors $Z_c$. However, the fake images are usually distributed between the specific classes (not exactly lie in the class distribution). The conditional latent vectors $Z_c$ are randomly located by applying mixture distribution on the class sample $c$. These obtained vectors are processed by the generator and the output will be fed into the discriminator of our proposed model. It is worth mentioning that, in our model; the embedding layer of the generator is randomly initialized and trained. However, to train such mixture distribution that could generate minority samples in the low-density region of the real data $\ell_G=(G,DC)$, we require a loss function to encourage the generator to do that. 
\begin{equation}
\ell_G = loss_{fm}
\label{eq:3}
\end{equation}
In this paper, we follow the work in [32] and apply feature matching function as a loss function in the generator. As proved in [32,37] Feature matching loss enrich the generator to generate samples from different data distribution but falls onto the margin, i.e., the generator tries to match the first-order features between the generator distribution and the true distribution $p_{td}$ of both components.
\begin{equation}
loss_{fm} = \min_{u_G}||E_{x\sim p_{td}(x)}[f(x)] - E_{z\sim p(z)} [f(G(z; u_G))]||
\label{eq:4}
\end{equation}

\subsection{Capsule Discriminative Model (DC)}
There have been a lot of works on how to address the imbalanced data learning problem. Our proposed model improves upon these methods by using a variation of GAN. The behavior of the convolutional networks for biased and imbalanced data has been extensively investigated [29,32,39], whereas the impact of capsule-net has received little attention so far. In this paper, we incorporate the capsule-net guideline [18] for the discriminator and used capsule-layers instead of convolutional layers, for the performance of imbalanced image classification. This way pushes the discriminator toward creating several clusters in its output embedding space. Each cluster corresponds to one of these vertices; belonging to the known set of classes or does not belong to any known classes (fake). The main differences between our discriminator and other GAN variations are; the way to generate the minority samples; stability and prevention of any changes in majority class space; and also, single output with several clusters. For a given feature vector $x$ with its corresponding groundtruth $y$, and a group of capsule vectors ${v_1,v_2,…,v_l }$ we train the model with $f(x)=||v_y||_2 > ||v_l||_2$. In fact, we need $||v_y||_2$ that should be larger than the other capsule vectors. If we assume $f(x)$  as a nonlinear vector function, and $w_k$ the weight vector for class $c$ as the standard-setting in [37], to train the discriminator-capsule we should minimize the $p_d (c|x)= \frac{exp(w_c^T f(x))}{\sum_{i=1}^{c+1}exp(w_i^T f(x))}$ . We define $\ell_D (G,DC)$ as below to train the discriminative model. Note that DC indicates our discriminator that uses a capsule-net.
\begin{equation}
\ell_{DC} = Ls + L_{pt}
\label{eq:5}
\end{equation}
The intuition behind the design of capsule discriminator is to improve the classification based imbalanced class images. Given that G generates minority samples from the mixture data distribution, we need to put forward some main condition for DC to guarantee the classification accuracies; (i) the samples from different components (class c) should be mapped to their corresponding components, (ii) the samples from different distribution which is not from any known components should be mapped into a particular cluster termed as fake samples, (iii) different clusters should be far enough. The most suitable way to satisfy the first and second conditions is margin loss since the discriminator of our model is based on a capsule-net. We modified the margin loss in the discriminator in order to keep the number of parameters low, to prevent harshly penalizing the generator hence causing the generator to fail in the training process. However, since we have c classes and we need to consider both sample classes, we modify the margin loss [24,40] for the discriminative model as: 
\begin{equation}
\begin{aligned}
h_c = DC(x_i) \\
Ls = \sum_{c\in C} T_c max(0, s^+ - ||h_c||)^2 \\
+\alpha(1-T_{c+1})max(0, ||h_c||- s^-)^2
\end{aligned}
\label{eq:6}
\end{equation}
DC indicated as discriminatory based capsule-net, $T_c$  is the target value, the output of discriminator will be one of $c$ class components which is shown by value zero, and $T_{c+1}$ is fake which is shown by value one; $h_c$  is the final output vector, and $x_i$  is the input vector which is received from the generator. According to [18], we set the parameters $(\alpha,  s^+ ,  s^- ) $ to 0.5, 0.9 and 0.1, respectively. For a classification task (discriminator D) involving $c$ classes, the final layer of the network contains $c$ results, each representing one class. Based on [18,40], the length of each capsule $||h_c||$ can be viewed as the probability of the image belonging to a particular class $c$. Furthermore, for condition (iii), we need to widen density gaps to help classification. Therefore, we used pull-away-term [38] to fulfill the condition. 
\begin{equation}
\begin{aligned}
l_{pt} = \frac{1}{m(m-1)}\sum_i \sum_{j\ne i}(\frac{x_i^T x_j}{||x_i||||x_j||})^2
\end{aligned}
\label{eq:7}
\end{equation} \par
where $x_i$ , $x_j$ are batch samples and $m$ is the batch size. It is worth mentioning that, in our model; the embedding layer of the generator is randomly initialized and trained. Finally, the proposed discriminator and generator will be optimized as: 

\begin{equation}
\begin{aligned}
D^* = max{(Ls(DC(x), T_c = 0)} + \\
max{Ls(DC(x), T_{c+1} = 1)} +{Eq. \ref{eq:7}}\\
then \rightarrow G^* = max{Ls(D(G(z)), T_c = 0)}
\end{aligned}
\end{equation}
Note that, here we eliminate $\alpha$ (down weighting factor) since in the generator model we do not use capsule. We have given the training details in Algorithm \ref{algo:1}. In step 1, the generator takes a noise vector z and class information $y_i$  as input with a uniform distribution and outputs the same format of original samples,  the generated samples $s_i^+ \&  s_i^- $ which can be either positive (minority) or negative (majority). In step 2, once the synthetic samples are generated, the classifier DC is designed to classify the input samples based on the known classes or unknown classes. In conclusion, the proposed algorithm solves the learning problem posed by imbalanced image classification by variation of GAN techniques. This approach can be outlined as follows: Divide the original samples into a training set and a test set; use the training set to train the GAN and tune its hyper-parameters; use the trained generator of the GAN to generate synthetic samples, and use the discriminator as a classifier to distinguish the results as fake or class $c$. 
\begin{algorithm}
\small{
 \caption{: training of our proposed method}
{\bf Input, } a set of minority and majority samples as $S^+,S^-$ respectively; \\
{\bf Output, } DC (assigning the samples as one of the above classes or fake);\\
{\bf Initialize, } $\omega$ and $u$ as the training parameters for discriminator and generators respectively; \\
$K = s^+ + s^-$, where $k$ is the number of samples including minority and majority class $(s^- > s^+) $ \\
{\it Train} the generator and discriminator with first $t$ iterations \\
{\it for} number of training, iterations {\bf do} \\
{\it sample} $i$ minority samples as $(x_1, x_2, ..., x_i )^{s+}$ \\
{\it sample} $i$ majority samples as $(\acute x_1, \acute x_2, ..., \acute x_i )^{s-}$ \\
{\it sample} $n$ noise as $(z_1, z_2,…, z_n)$ \\ 
{\it calculate} the number of samples to be generated $n_+ = integer(dist_+ \times (s^- - s^+)\times \varphi)$ \\
${\bf (s^- -s^+)}$ is  the difference between the number of majority samples and the minority samples, $\varphi$  is the level of balance to be generated (i.e., value 1, means 100\%, and 0.5 meaning 50\%) \\
{\it generate} samples $\hat x \leftarrow G(z, y_i)$ \\
{\it update} the discriminator by: $\omega \leftarrow \Delta_\omega \frac{1}{i}\sum_{c\in C} Ls + l_{pt} $ \\
{\bf end for} the number of training iterations(t steps) {\bf do} \\
{\it sample} $i$ minority samples as $(x_1, x_2, ..., x_i )^{s^+}$ \\
{\it sample} $i$ majority samples as $(\acute x_1, \acute x_2, ..., \acute x_i )^{s^- }$ \\
{\it update} the generator by: $u\leftarrow \Delta_u \frac{1}{i}\sum_i l_{fm}$ \\
{\bf end} \\
{\bf until} convergence \\
\label{algo:1}
}
\end{algorithm}\par
To summarize, we proposed a new variation of GAN for imbalanced data learning, to generate minority samples in a low-density area of the true data by using feature matching loss that could enrich the generator distribution and using capsule-discriminator to create several clusters in its output space. In this proposed way, we can provide better meaning to the discriminator that, not only distinguishes between fake and real images but also identify the corresponding class for each input. 
\begin{table*}[h]
\centering
\caption{Results of discriminator based on the capsule and convolutional nets on MNIST, CIFAR-10, CelebA, based on the error rate}
\begin{tabular}{l| c c c| c c c| c c c}
\hline
Datasets & \multicolumn{3}{c}{MNIST} & \multicolumn{3}{c}{CIFAR-10} & \multicolumn{3}{c}{CelebA} \\ \cline{2-10} 
\hline\hline
No. of samples (n) & 100 & 500 & 1000 & 100 & 500 & 1000	& 100 & 500 & 1000  \\  [0ex]
Caps-discriminator & 0.8153 & 0.7329 & 0.6953 & 0.2408 & 0.2197 & 0.0698 &	0.4169 & 0.3418 & 0.0940 \\   [0ex]
Conv-discriminator & 0.8204 & 0.7927 & 0.7536 & 0.2563 & 0.2300 & 0.1150 &	0.5721 & 0.4839 & 0.1007 \\ [0ex]
\hline
\end{tabular}
\label{tab:1}
\end{table*} \par

\subsection{Implementation details}
We have incorporated the capsule-net in GAN to learn from imbalanced data. Similar to DCGAN [15], we used convention CNN in designing the generator, however, motivated by [18,34], we designed our discriminator based on capsule-net. The main idea behind this implementation is that, in the proposed CapsAN, the generator acts as the oversampling, and the discriminator is our classifier. To implement the DC (capsule-discriminator), we followed the original paper [18]. However, the final layer of our discriminator has a $1\times16$ dimensional capsule, instead of having $10\times16$. This is the discriminator’s output which represents, whether the input image is fake or belonging to one of the corresponding classes. We initialize the model with a $5\times5$ convolutional layer. There are 32 primary capsules with $4\times4$ pose linear transformation. The primary capsules are followed by another $5\times5$ convolutional layer. The final capsule is the output layer with 16 dimensions. All convolutional layers are composed of kernels of size $5\times5$ with a stride 2 followed by spectral normalization [49]. For each of the 16-dimensional vectors, we compute the margin loss in Eq. \ref{eg:6}. The first layer in G is the latent samples, which is followed by a fully connected layer. In addition, the generator involves 5 fractional-convolutional layers (instead of transpose convolution) which are responsible for upsampling a $512\times512$ dimensional vector. All fractional convolutional layers are composed of the kernel of size $3\times3$ with stride 1 followed by conditional batch normalization [50]. The earlier methods based GANs in order to ease the training difficulties suggested using batch normalization in both generator and discriminator. However, these methods usually adapted to eliminate covariate shifts. In this paper, we used a different setting for the discriminator and substituted the SN (spectral normalization) [49] with the batch normalization which helps deal with the training problem. Moreover, we thoroughly set the weight initialization by using trivial $N (0, 1)$ and explicitly scaled the weights as: $W_l=\frac{w_i}{c}$, where $c$ is the normalization in each layer $l$ and $w$ is the weight; this initializer scenario is proved by He et al. [7] which helps to be independent of the scale parameter during the training. To control the magnitude in the generator we normalize the feature vector in the generator after each fractional convolutional layer. We used \verb|LeakyReLU| [51] activation function with a leaky slop of 0.2 in the generator and discriminator for all layers, except for the final layer. We observed the 0.001 learning rate is too high, thus we set to 0.0002. Also, the momentum term with the standard value of 0.9 does not match with our parameters and we set the Adam optimizer [12] with value 0.5 instead. The proposed networks were trained on a workstation equipped with an Intel i7-6850K CPU with a 64 GB Ram and an NVIDIA GTX Geforce 1080 Ti GPU and the operating system is Ubuntu 16.04. We used Keras 2.1.2, the deep learning open-source library and TensorFlow 1.3.0 GPU as a backend deep learning engine. Python 3.6 is used for all the implementations.  The baselines were implemented using their source codes.

\section{Empirical analysis and discussion}
\label{sec:5}
We now present our main contribution of the proposed model that can effectively tackle the imbalance problem in real-world images. The proposed model improves learning form the imbalanced data by exploiting all useful information from minority and majority class samples in the generation of new samples. The generated minority samples are handled in a way to prevent overfitting and outlier production. We validate the proposed model on the most popular and recognized image datasets, namely CIFAR10 [45], CelebA [47], Fashion-MNIST [46], GTSRB [48], and MNIST [44]. CIFAR-10 is well known and used for a wide range of applications; it contains 60,000 images from a tiny image dataset with 10 different classes. MNIST (mixed national institute of standard and technology) is an example of the imbalanced image that contains a large database of 70,000 digits (0 to 9). GTSRB is a traffic sign recognition dataset with 51,840 images in 43 classes. All the selected datasets are already imbalanced; we randomly select one class from each dataset and follow [5,39] to induce an imbalance rate of 40\%, 20\%, 10\%, 5\% and 2.5\% in them. We label this imbalanced data as Data 1, Data 2, Data 3, Data 4, and Data 5 respectively. In order to create a highly imbalanced data, we randomly select a class and drop a significant amount of its instances from the training set. Since the aim of this paper is binary classification, we repeat this process for every two classes and train the proposed models and the baselines for each obtained imbalanced dataset. GTSRB dataset is imbalanced; we do not further imbalance it. The following results shown are averaged over 10 runs. 
\begin{figure*}
\centering
 \includegraphics[width=0.93\textwidth]{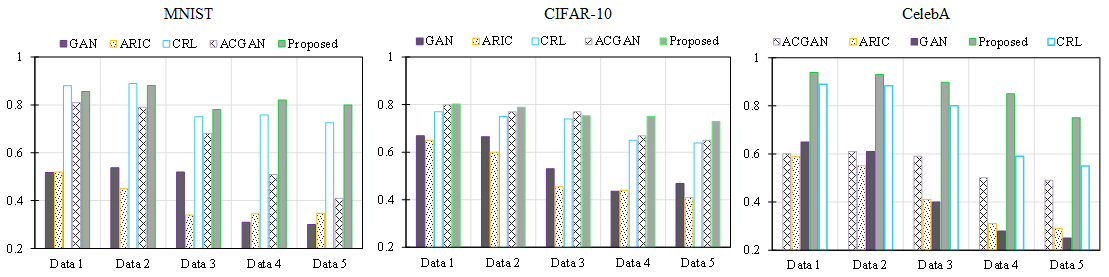}
\caption{Balanced Accuracy with number of imbalance percentage; 40\%, 20\%, 10\%, 5\%, 2.5\% corresponds to Data1, Data2, Data3, Data4, Data5, Data6 and Data7 respectively.  Higher values are better results}
\label{fig:3}       
\end{figure*}

\begin{figure*}
\centering
 \includegraphics[width=0.95\textwidth]{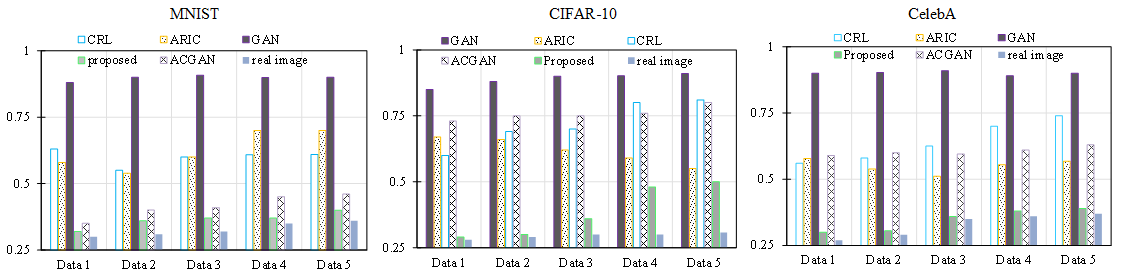}
\caption{(SSIM)- Similarity with number of imbalance percentage; 40\%, 20\%, 10\%, 5\%, 2.5\% corresponds to Data1, Data2, Data3, Data4, Data5, Data6 and Data7 respectively. Lower values are better result.}
\label{fig:4}       
\end{figure*}

\subsection{Quantitative assessments }
The quality of the generated samples will be assessed based on some criteria. The newly generated samples must be from the target class (with no overlapping) and not be repetitive; the newly generated samples must be different from the real ones in the training set. Failing to meet this criterion may degrade the quality of the generated images. However, in order to show the impact of capsule-net in GAN comparing to convolutional GAN, we follow the work by Im et al. [33] which introduced new adversarial metrics for GAN models by placing each generator against the opponent’s discriminator. It means that, if we assume our proposed model as $S_2$ and the regular GAN as $S_1$, then $S_1= (G_1, D_1), also, S_2= (G_2,D_2)$;$ G_1$, engage against $D_2$, $G_2$ engage against $D_1$. We followed their implementation \footnote{\fontfamily{pcr}\selectfont{https://github.com/jiwoongim/GRAN/battle.py}} , we calculate the ratio of classification accuracy for real test data and generate samples as:$ r_{generaated}=\frac{AC(D_1 (G_2 (z)))}{AC(D_2 (G_1 (z)))}$ ; $r_{real}=\frac{AC(D_1 (\acute x))}{AC(D_2(\acute x))}$, $\acute x$ indicates the test data samples and $AC$ is the accuracy. To have a better result for the Capsule-GAN the above equations must be satisfied by $r_{generaated}<1 $ and $r_{real}=1$. In our experiment based on CIFAR-10, CelebA , and MNIST datasets, we achieved $ r_{generaated}=0.83 , 0.78$ and $r_{real}=1.0, 1.0 $ respectively. Therefore, the results convey that, capsule GAN win against Conv-GAN. \par
We also use the {\fontfamily{lmtt}\selectfont{scikit-learn}} \footnote{\fontfamily{pcr}\selectfont{http://scikit-learn.org/}} package to randomly generate 1,000 images using our proposed model and regular GAN (Conv-GAN). Three set samples have been used as  $n\in (100,500,1000)$. Table \ref{tab:1} is evaluated on MNIST and CIFAR-10 datasets. The result is the example to show that discriminator by adopting a capsule-net consistently performs better than convolutional GAN for both the image datasets. The error rate is high since we use raw pixels for the classification in order to have a fair evaluation. Our proposed model compares to convention GAN is capable to generate minority samples with higher variety which leading to better classification performance. We used the following metrics [52, 53] to verify that the generated images by our model are representative of the target classes and also the predicted classes match the target class. 
$BA=\frac{\sum_{c=1}^c n_{cc}}{\sum_{c=1}^c, d=1^n_{cd}}$ ; $F_M = \frac{2\sum_{c=1}^c R_c \sum_{c=1}^c P_c}{c\sum_{c=1}^c R_c \sum_{c=1}^c P_c}$ ; $R_c = \frac{n_{cc}}{\sum_{d=1}^c n_{cd}}$ ; and $P_c = \frac{n_{cc}}{\sum_{d=1}^c n_{cd}}$; where $c$ is the number of class ; $n_{cc}$  and $n_{cd}$ denote the number of class samples that are correctly predicted as class c and incorrectly predicted as class $d$, respectively; $R_c$ is the recall term and $P_c$ indicates the precision of class $c$. 

\subsection{Accuracy and Variability of the generated image}
In order to assess the quality of generated images, we design the discriminator by using a capsule-net instead of the convolutional neural network. The methodology for evaluation is as follows: Firstly, we need to create an imbalanced dataset; thus for each class, we remove some part of that class (dropping a percentage of images for each class from the training set) then we create imbalanced dataset by a percentage of 40\%, 20\%, 10\%, 5\% and 2.5\%. By using our generator, we generate minority samples from novel multivariate distribution; train the designed discriminator for the balanced datasets and measure the balanced accuracy over the testing set. The result is given in Figure \ref{fig:3}. We observed the regular GAN has the worst accuracy compared to other methods. Our proposed model returns better results in most of the cases for MNIST, CelebA, and CIFAR-10 datasets. In particular, for severe imbalance ratio (10\%, 5\%, and 2.5\%) our model is capable to generate high variety samples that the capsule-discriminator can classify with the highest accuracy. However, a strong imbalancedness ratio can deteriorate the quality of generated samples which results in a low accuracy, this is most evident for GAN [11] and ARIC [25] on the MNIST and CelebA datasets. We observe again that, for the small imbalance percentage, the best result is achieved by ACGAN [20] and CRL [7] approaches. This implies that, as the imbalance level becomes severe, in practice from Data3 to Data5, there is a fall in the value of accuracy; thus, the more severe imbalance the less operative the resampling. Furthermore, in order to show the variability of the generated images, we used the SSIM metric [41] to measure the similarity between two images. The result is a decimal value between 0 and 1. It returns value 1 if the set of images has little variability and different from the original one, whilst, 0 value means, the images are identical and thus the quality is well preserved. This diversity result shows in Figure \ref{fig:4}. Note that we get a distinct SSIM value for each class, and then, average these values per sample SSIMs. We also include the real image value in our evaluation. Generally, the real images have more variables than the generated version (lower SSIM). We used the real image value as a reference and analyzed the other results based on this value. From the results, it observes that our proposed model exhibits good variability and shows low SSIM, almost near to the real image values. However, GAN gives the best SSIM which is very close to 1. It means that the generated image by GAN has little variability and different from the real one. 

\begin{table}[h]
\small{
\centering
\caption{FIDs on different datasets from different methods. Lower FID represents higher quality for generated images. The lowest values are better results.}
\begin{tabular}{p{.12\textwidth}|p{.05\textwidth}p{.06\textwidth}p{.05\textwidth}p{.09\textwidth}}
\hline
GAN variations & MNIST & CIFAR-10 & CelebA & FASHION-MNIST \\ 
\hline\hline
{\bf{Real image} \protect \footnotemark} & 1.25 & 5.19 & 2.27 & 2.60    \\  [0ex]
GAMO [30] & 7.53 & 14.59 & 12.25 & 11.47  \\   [0ex]
GAN [11] & 17.81 &	37.73 & 41.27 & 30.39 \\ [0ex]
DAGAN [42] & 13.29 & 31.48 & 44.98 & 36.10 \\
{\bf CapsAN} & 4.81 & 9.07 & 13.06 & 10.23 \\
\hline
\end{tabular}
\label{tab:2}
}
\end{table} 
\footnotetext{\fontfamily{pcr}\selectfont{the result is taken from its original paper: arXiv preprint arXiv:1711.10337 }}

\subsection{Fr\'echet Inception Distance}
\begin{figure*}
\centering
 \includegraphics[width=0.89\textwidth]{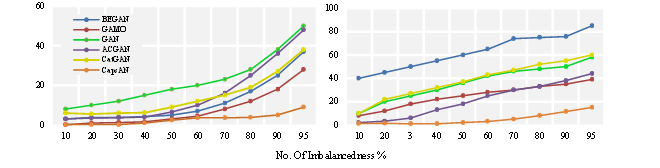}
\caption{Comparison of FID across different imbalancedness percentage. The lower value gives better performance. Left: MNIST data, Right: CIFAR-10. }
\label{fig:5}       
\end{figure*}

We used Fr\'echet Inception Distance (FID) [35] to quantify the quality of the generated image. The results evaluated with (FID) can be found in Table \ref{tab:2}. These results are obtained by using GAN variations on different datasets to demonstrate the abilities to generate a high-quality image. This is confirmed by the lower FID value between the real image and its corresponding generated one [43]. We used real image values as a reference and analyzed our performances based on this value. The real image values are obtained from . As it observes, our proposed model achieves the lowest (best) FID among all baselines on MNIST and CIFAR10 dataset. GAMO [30] gives better FID for CelebA and Fashion-MNIST dataset. We also evaluate the quality of the generated image, in Figure \ref{fig:5}, in particular, to show the impact of using mixture distribution for drawing the minority-class samples. We follow the structure in the original work [35] and use the pre-trained Inception-v3 model. We used the same number of generated samples as the training set. We compare all the methods by using different imbalanced percentage from 10\% to 95\%. It can be inferred that, in both the experiments, by increasing the number of the imbalanced percentages still ensuring the stable training processes for our proposed model. This leads us to say that, our model consistently outpaces other methods in all cases, in particular for a larger number of imbalanced percentages. Other techniques perform better in low imbalance ratio; this is most evident for ACGAN [20] and GAMO [30]. We also found that, for MNIST dataset, the baselines to be quite unstable and sensitive to the number of imbalance ratios. As the level of imbalance become more from 50\% to 95\%, their values also gradually are increased. 

\subsection{Quality of the proposed classifier based on capsule discriminator}
\begin{figure*}
\centering
 \includegraphics[width=0.8\textwidth]{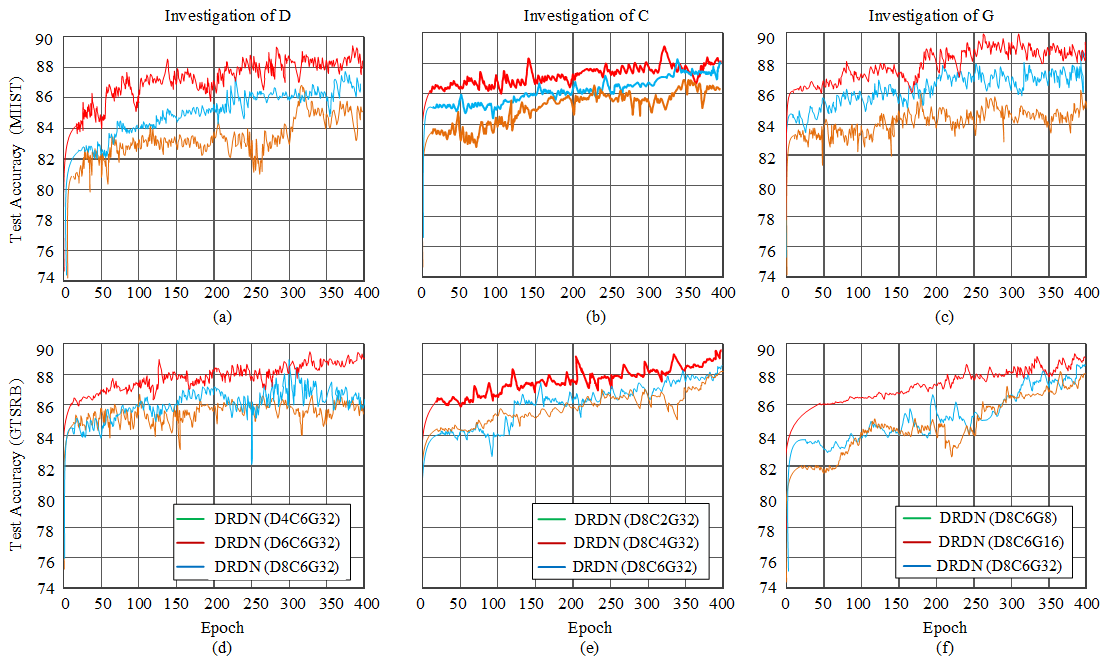}
\caption{Convergence investigation of the proposed generator on two datasets with different values of C, P, and BN.}
\label{fig:6}       
\end{figure*}

We evaluate the effect of our proposed generative model based on different metrics. The results are given in Figure \ref{fig:6}. Here, we study the basic network parameters on MNIST and GTSRB datasets: The number of Convolution layers (denoted as C), the number of Max-pooling layers (denoted as P), and the number of conditional batch normalization layers (denoted as BN). As presented in Figures \ref{fig:6}(a), (b), (d) and (e), larger C or P may boost up the performance. This is generally because the network goes deeper with larger C or P. As our proposed model supports larger BN, we also realize larger BN Figure \ref{fig:6}(c) and Figure\ref{fig:6}(f) led to superior performance. Alternatively, the proposed model with smaller C, P, or BN has lower testing performance. 

\begin{figure*}
\centering
 \includegraphics[width=0.8\textwidth]{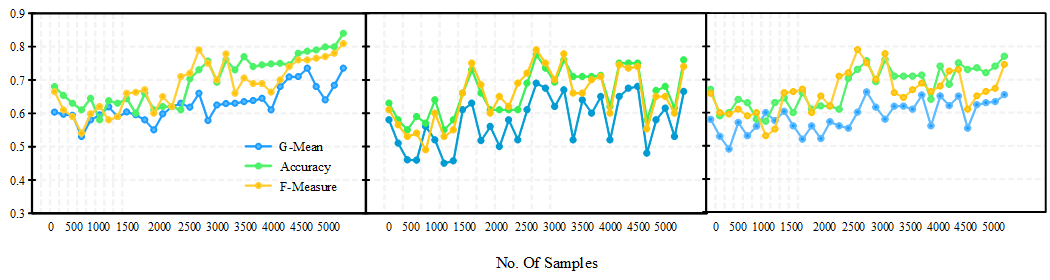}
\caption{Performance of discriminator over different training sizes. Left to right: MNIST, CIFAR10, and CelebA dataset. High recall and precision imply the best performance; high precision but low recall implies that there is a lack in diversity; low precision but high recall implies that the model can decently reproduce the sharp image but fails to capture convexity; low precision and low recall implies an undesired performance.}
\label{fig:7}       
\end{figure*}

\begin{figure}
 \includegraphics[width=0.45\textwidth]{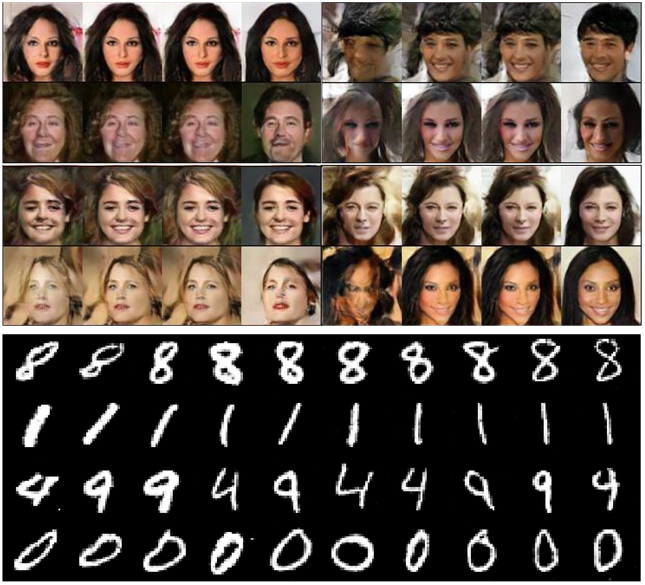}
\caption{The generated samples by the proposed model.}
\label{fig:8}       
\end{figure}
We also assess the performance of our proposed model across the test benches. In Figure \ref{fig:7}, we evaluate the effect of the generated sample size on the performance of the discriminator. To this end, we used a different number of generated samples as training and then used the real images to test the three metrics of accuracy, G-mean, and F-measure. The results imply that as the training size increases, the accuracy also increases in all three image datasets. Similarly, when the number of generated samples increases, the F-measure is basically consistent with the accuracy. The G-mean values are slightly lower than accuracy in all cases but it tries to be consistent overall. 

\begin{figure*}
\centering
 \includegraphics[width=0.65\textwidth]{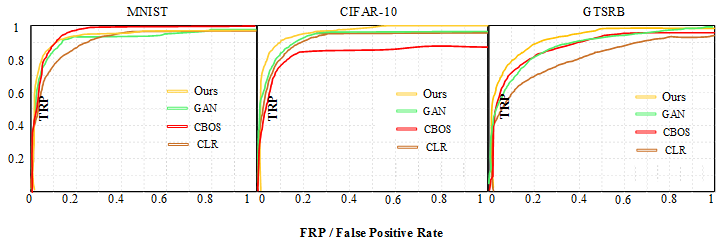}
\caption{From left to right, ROC curves of training a model upon different image datasets. The baselines are CRL [7], CBOS [5] and GAN [11]. True Positive Rate (TPR) value starts from 0 to 1 and False Positive Rate (FPR) value starts from 0 to 1.}
\label{fig:9}       
\end{figure*}

Figure \ref{fig:8} shows CapsAN generated samples. It clearly observes that our model can generate more realistic and diverse images, which is confirmed by the lower Fr\'echet Inception Distance as shown in Figure \ref{fig:5} and Table \ref{tab:2}. We provide a fair assessment of the different resampling techniques using recall, precision, and f-measure in Table \ref{tab:3}. A high precision score implies that the generated samples from the model distribution are close to the real data distribution. Similarly, if the generator generates something close to any sample from all available data distribution, its recall is high. A good generative model should be able to capture all the available variations and generate the training set [2,32]. The results are averaged over 10 runs. We first, created imbalanced dataset from MNIST dataset, with different imbalance ratio as: 50\%, 60\%, 70\%, 80\%, 90\%, 95\% and 98\%. These datasets are labeled as Data1, Data2, Data3…, and Data7 respectively. The way to create imbalanced data is explained in section 5.1; i.e. Data 1 contains 50\% imbalance and Data7, contains 2\% imbalance. We also included the results from the original imbalanced image (without using any resampling technique); the results clearly show that all the techniques improve upon the original imbalanced image. However, the original image has a good performance on Recall. This is due to the fact that using standard GAN on the imbalanced original image mostly prone to the majority class samples and this may increase the results in the good classification of majority samples. Also, we can observe that as the level of imbalance increases (Data4 to Data7); the results are decreases mostly for all the metrics. This leads us to say that, severe imbalance percentage deteriorates the effect of resampling technique. Based on the results in (Data1 to Data3), we see that the Recall value of all the techniques sans our model decreases as compared to the original imbalanced data. This conveys that, our proposed model doesn’t take the majority class space into consideration when generating the minority class, while, the other resampling techniques in addition to the minority class space, have an effect on the majority class space. It is also interesting to note that SMOTE [9,28] despite being an effective oversampling technique, but have not received much attention in deep learning. \par

The reason is, the feature extraction and classification in deep learning techniques are in an end to end process, thus, it is difficult to incorporate oversampling similar to SMOTE in such frameworks. This comparison shows with respect to the other techniques- our model generally outstrips the baselines. The other techniques give better performance in a small imbalanced ratio; this is evident for CLR and ADASYN. The best performances of each metric are marked in blue. The results lead us to confidently say that the other resampling techniques which are used in our experiments do not efficiently handle severe imbalanced datasets and only perform better in (Data1 to Data3). We can conclude that our model is superior to other methods of generating a high variety of minority class images. In addition, discriminator using a capsule-net can be applied to improve the final accuracy for the data with the multivariate distribution. Next, we plot the averaged ROC curves in Figure \ref{fig:9}. We showed that the proposed model outperforms the baselines in MNIST and GTSRB datasets; however, in the CIFAR-10 dataset, the CRL method performs better than other methods.   

\begin{table*}
\centering
\small{
\caption{Comparing the performance of resampling techniques with various evaluation metrics. Note that, in the table, imbalanced refers to the standard GAN methods without using any resampling technique. The best results are \textcolor{blue}{highlighted}.}
\begin{tabular}{c| p{3cm}p{2cm}p{2cm}p{2cm}p{2cm}p{2cm}}
\hline
\multirow{4}{*} & Methods & Precision & Recall & Accuracy & F-measure & G-mean \\
\hline\hline
\multirow{4}{3em}{Data 1} & Imbalanced [11] & 0.659 & 0.916 & 0.803 & 0.756 & 0.769 \\ [-0.5ex]
                                      & SMOTE [28] & 0.812 & 0.873 & 0.809 & 0.816	& 0.814 \\ [-0.5ex]
                                       & ADASYN [8] & 0.893 & 0.887 & 0.878 & 0.917 & \textcolor{blue}{0.923}  \\ [-0.5ex]
                                      & CRL [7] & 0.919 & 0.891 & 0.884	 & 0.912 & 0.903 \\ [-0.5ex]
                                      & CapsAN & \textcolor{blue} {0.927} &\textcolor{blue}{ 0.921} & \textcolor{blue}{0.889} & \textcolor{blue}{0.921} & 0.915  \\ [-0.5ex] \cline{1-7} 
\multirow{4}{3em}{Data 2} & Imbalanced [11]	& 0.651 &	0.908 &	0.785 &	0.756 &	0.734 \\ [-0.5ex]
& SMOTE [28] &	0.788 &	0.817 &	0.798 &	0.803 &	0.809 \\ [-0.5ex]
& ADASYN [8] &	0.873	& 0.869 &	0.881	& 0.879 &	0.883 \\ [-0.5ex]
& CRL [7] &	0.915 &	0.897	& \textcolor{blue}{0.903}	& \textcolor{blue}{0.910}	& 0.898 \\ [-0.5ex]
& CapsAN &	\textcolor{blue}{0.921} & \textcolor{blue}{0.918} &	0.891 &	0.905	& \textcolor{blue}{0.911} \\  [-0.5ex] \cline{1-7} 
\multirow{4}{3em}{Data 3} &	Imbalanced [11] &	0.637 &	0.915	& 0.803 & 0.744 &	0.721 \\ [-0.5ex]
	& SMOTE [28] &	0.783 & 	0.833 &	0.799	& 0.811 &	0.802 \\ [-0.5ex]
	& ADASYN [8] &	0.871	& 0.893 &	0.871 &	0.899 &	0.887 \\ [-0.5ex]
	& CRL [7] &	\textcolor{blue}{0.904}	& 0.908 &	\textcolor{blue}{0.884}	& 0.915	& \textcolor{blue}{0.919} \\ [-0.5ex]
	& CapsAN &	\textcolor{blue}{0.909}	& \textcolor{blue}{0.935} &	0.873	& 0.929 &	0.903 \\  [-0.5ex] \cline{1-7} 
\multirow{4}{3em}{Data 4} &	Imbalanced [11] &	0.529 &	0.893	& 0.754 &	0.769 &	0.716 \\ [-0.5ex] 
	& SMOTE [28] &	0.779	& 0.802	& 0.787	& 0.796	& 0.799 \\ [-0.5ex] 
	& ADASYN [8] &	0.882	& 0.889 &	0.875	& 0.886 &	\textcolor{blue}{0.891} \\ [-0.5ex]
	& CRL [7] &	0.893 &	\textcolor{blue}{0.923}	& \textcolor{blue}{0.891} &	0.879	& 0.874 \\ [-0.5ex]
	& CapsAN &	\textcolor{blue}{0.895}	& 0.914 &	0.886	& \textcolor{blue}{0.914} &	0.882  \\ [-0.5ex] \cline{1-7} 
\multirow{4}{3em}{Data 5} &	Imbalanced [11]&	0.484&	0.723&	0.715&	0.709&	0.693 \\ [-0.5ex]
	&SMOTE [28]&	0.736&	0.748&	0.751&	0.743&	0.758\\ [-0.5ex]
	&ADASYN [8]&	0.876&	\textcolor{blue}{0.893}&	0.865&	0.873&	\textcolor{blue}{0.889} \\ [-0.5ex]
	&CRL [7]&	0.881&	0.878&	\textcolor{blue}{0.872}&	0.875&	0.881 \\ [-0.5ex] 
	&CapsAN &	\textcolor{blue}{0.883}&	0.881&	\textcolor{blue}{0.874}&	\textcolor{blue}{0.889}&	0.886\\  [-0.5ex] \cline{1-7} 
\multirow{4}{3em}{Data 6} &	Imbalanced [11]&	0.407&	0.689&	0.637&	0.659&	0.647 \\ [-0.5ex]
	&SMOTE [28]&	0.689&	0.665&	0.673&	0.669&	0.661 \\ [-0.5ex]
	&ADASYN [8]&	0.804&	0.815&	0.793&	0.786&	0.795 \\ [-0.5ex]
	&CRL [7]&	0.826&	0.853&	0.849&	0.823&	0.819 \\ [-0.5ex]
	&CapsAN &	\textcolor{blue}{0.847}&	\textcolor{blue}{0.869}&	\textcolor{blue}{0.857}&	\textcolor{blue}{0.836}&	\textcolor{blue}{0.845} \\ [-0.5ex] \cline{1-7}				
\multirow{4}{3em}{Data 7} &	Imbalanced [11]&	0.418	&0.625&	0.603	&0.594&	0.618 \\ [-0.5ex]
	&SMOTE [28]&	0.613&	0.609&	0.618&	0.597	&0.611 \\ [-0.5ex]
	&ADASYN [8]&	0.779&	0.806&	0.783	&0.774&	0.765 \\ [-0.5ex]
	&CRL [7]&	0.806&	\textcolor{blue}{0.844}&	\textcolor{blue}{0.811}&	0.795&	0.784 \\ [-0.5ex]
	&CapsAN&	\textcolor{blue}{ 0.812}	& \textcolor{blue}{0.849}&	 \textcolor{blue}{0.819}&\textcolor{blue}{0.807}&	\textcolor{blue}{0.823} \\ [-0.5ex]
\hline
\end{tabular}
\label{tab:3}
}
\end{table*} 

\subsection{Discussion}
The proposed model is trained to properly classify the samples from all the classes. The strength of our model lies in the following components; generating minority samples with a high variety from the mixture data distribution; train the generator using feature matching function; incorporate capsule-net into the discriminator (DC); train the multiclass discriminator to create several clusters in its output space, the output of DC will be as the input belongs to the known set classes $(c_1, c_2,…, c_n)$, or Fake. Most of the imbalance problems are dominated by binary classifications, but our model is based on multiclass discriminator. However, we used our datasets for a two-class classification problem but the proposed model can also be applied for the multiclass dataset. We conducted several experiments. 1) Evaluate the quality and variability of the generated images. 2) Evaluate the quality of the capsule-discriminator as a classifier. 3) Overall performance of the proposed model. More imbalanced datasets including a medical dataset, remote sensing dataset can be used to test the efficacy of the proposed model. We have shown that a generator that is trained with feature matching loss has an impressive ability to handle the mixture data distribution. Indeed, the performance of the generative model (quality of the generated samples) strongly lies in the problem of computing the distance to the multivariate samples. In the case where a generator generates samples from a mixture of data distribution (minority and majority class), we showed that the capsule-discriminator can be a powerful multiclass classifier. When the multiclass discriminator classifies a sample as “fake”, it conveys that, most likely this sample is not belonging to any known classes; otherwise the class with high probability will be assigned to the sample.  In the future, we would like to search for new loss functions for the generators that will enrich the generator distribution and will improve the classification performance. 

\section{Conclusion}
\label{sec:6}
A lot of works have emerged focusing on class imbalance problems; however, these works mostly affect the majority class space or even did not consider the majority samples in data generation. In this work, we proposed a novel GAN variation called CapsAN, for handling the class imbalanced problems. To achieve the results, we follow a new mechanism: Motivated by the success of capsule-net in complex data, we used its advantages for designing our discriminator. We incorporate capsule-net into the GAN framework to improve the performance of the discriminator and in addition alleviates the overfitting problem. The discriminator of our model creates a single output with several clusters in its embedding space. This leads the generator to generate samples with more variety. Our generative model uses mixture distance distributions between the majority and minority class to infer the number of new samples for the minority class while it does not affect learning from the majority space. The generative model with the mixture distribution is trained by the Feature matching loss and it can prevent the generation of an outlier. The central differences between our model and other relative techniques like, GAMO, ACGAN are: the way to generate minority class, incorporating spatial structure of the majority class in the generated samples; the way to select how many samples to be generated; using capsule-net instead of convolutional network to make the model more stable and prevent of overfitting. We believe that incorporating the capsule-net into the discriminator, makes a powerful classifier in multiclass imbalanced learning as well.

\end{document}